\documentclass[conference]{IEEEtran}
\IEEEoverridecommandlockouts
\usepackage{cite}
\usepackage{amsmath,amssymb,amsfonts}
\usepackage{algorithmic}
\usepackage{graphicx}
\usepackage{textcomp}
\usepackage{xcolor}
\def\BibTeX{{\rm B\kern-.05em{\sc i\kern-.025em b}\kern-.08em
    T\kern-.1667em\lower.7ex\hbox{E}\kern-.125emX}}
\begin{document}

\title{Hybrid Machine Learning Model for Detecting Bangla Smishing Text Using BERT and Character-Level CNN\\}

\author{\IEEEauthorblockN{Gazi Tanbhir}
\IEEEauthorblockA{\textit{\small Department of Computer Science and Engineering} \\
\textit{\small World University of Bangladesh}\\
Dhaka - 1230, Bangladesh \\
gazitanbhir@gmail.com}
\and
\IEEEauthorblockN{Md. Farhan Shahriyar}
\IEEEauthorblockA{\textit{ \small Department of Computer Science and Engineering} \\
\textit{\small World University of Bangladesh}\\
Dhaka - 1230, Bangladesh \\
farhanshahriyar.cse@gmail.com}
\and
\IEEEauthorblockN{Khandker Shahed}
\IEEEauthorblockA{\textit{\footnotesize Department of Electrical and Electronics Engineering} \\
\textit{\small Jashore University of Science and Technology}\\
Jashore – 7408, Bangladesh \\
khandkershahed23@gmail.com}
\and
\IEEEauthorblockN{Abdullah Md Raihan Chy}
\IEEEauthorblockA{\textit{\footnotesize Department of Computer Science and Engineering} \\
\textit{\small Southern University Bangladesh}\\
Chattogram,Bangladesh \\
raihanchowdhuryengineer@gmail.com}
\and
\IEEEauthorblockN{Md Al Adnan}
\IEEEauthorblockA{\textit{\footnotesize Institute of Information Technology} \\
\textit{\small Noakhali Science and Technology University}\\
Noakhali-3814, Bangladesh \\
md.al.adnan00@gmail.com	}
}

\maketitle

\begin{abstract}
Smishing is a social engineering attack using SMS containing malicious content to deceive individuals into disclosing sensitive information or transferring money to cybercriminals. Smishing attacks have surged by 328\%, posing a major threat to mobile users, with losses exceeding \$54.2 million in 2019, yet the issue remains significantly under-addressed despite its growing prevalence. This paper presents a novel hybrid machine learning model for detecting Bangla smishing texts, combining Bidirectional Encoder Representations from Transformers (BERT) with Convolutional Neural Networks (CNNs) for enhanced character-level analysis.

Our model addresses multi-class classification by distinguishing between Normal, Promotional, and Smishing SMS. Unlike traditional binary classification methods, our approach integrates BERT’s contextual embeddings with CNN’s character-level features, improving detection accuracy. Enhanced by an attention mechanism, the model effectively prioritizes crucial text segments. Our model achieves 98.47\% accuracy, outperforming traditional classifiers, with high precision and recall in Smishing detection, and strong performance across all categories. 
\end{abstract}


\section{Introduction}
Smishing is a kind of social engineering assault in which cyber-criminals send fraudulent text messages to trick individuals into downloading malicious software, revealing confidential information, or transferring money. Combining SMS and phishing is where the name "smishing" originates\cite{ibm_what}. The financial consequences of malicious cyber activity have increased by around 150\% in recent years~\cite{a2023_mace}. Furthermore, smishing, among the most widespread instances of social engineering, has increased by a surprising 328\%~\cite{laudon_2020_security}. As stated in the FBI's report on Internet Crimes for 2020, smishing losses in 2019 totaled more than \$54.2 million. In spite of the 95\% increase in mobile customers over the past few decades~\cite{a2014_number}, it is alarming that this threat is still largely neglected and presents significant risks. Seven out of ten persons claim to have received a questionable text message, indicating that scams are more frequently attempted via text messages \cite{_45}.

Individuals' mobile devices are vulnerable to attack by ransomware, malware, clickjacking, viruses, Trojan horses, and other risks due to smishing~\cite{kaspersky_2021_sms}. Attackers could potentially be capable of obtaining sensitive personal data, including location data, banking credentials, identification numbers, and passwords, as a result of these criminal activities~\cite{thomas_2017_data}. If this personally identifiable information (PII) is disclosed without permission or awareness, it might result in serious fraudulent activities~\cite{milne_2016_information, mdlutforrahman_2023_users}.

Compared with email-based marketing, SMS-based marketing causes faster response times and attracts five times more engagement~\cite{messentecommunicationsltd_2023_sms}. Additionally, a study~\cite{mdlutforrahman_2023_users} showed how vulnerable mobile users remain to smishing attacks. Because of these benefits, users of mobile devices have a threefold higher likelihood compared to desktop users to engage attackers in web-based phishing attacks~\cite{alkhalil_2021_phishing}. People are led to assume that SMS messages are real by their altered text and phishing URLs, which are frequently designed to look like reputable organizations such as banks or police enforcement. This strategy uses tiny modifications, including misspellings or using names that are similar but slightly different, to draw attention away from suspicion and toward the reliability of the SMS~\cite{milne_2016_information, mdlutforrahman_2023_users}.

This is difficult to differentiate between spam and smishing messages, a domain that has previously been addressed for the classification of smishing and spam~\cite{mohdfoozy_2014_a}. This is due to the prior rule-based categorization of ham and spam messages, as recommended by Oluwatobi Noah, et al.~\cite{akande_2022_development} and Jain, et al.~\cite{jain_2018_rulebased}.

Recent research, such as that cited in~\cite{khan_2023_exploring} and~\cite{sultana_2023_bilingual}, is centered on the detection of text spam in Bangla, mostly utilizing binary classification techniques that make use of BERT and other deep learning models. These approaches have been successful in classifying spam and non-spam messaging, but they have not yet addressed the challenges of multiclass classification, which is necessary for advanced detection tasks like classifying different types of smishing SMS.

To the best of our knowledge, the implementation of promotional messages in Bangla datasets has been mainly ignored by previous studies. This absence is crucial because it increases the possibility of improper categorization whereby the training data's inability to distinguish between spam and genuine promotional messages could lead to misinterpretation. The models might find it troublesome to differentiate between distinct kinds of messages in the absence of clear labeling for promotional content, which could lower the overall effectiveness of spam detection systems.

Our study is the first to apply multiclass classification to detect smishing text in the Bangla language. We propose a unique approach that uses deep learning techniques to integrate BERT's attention mechanisms with hybrid feature extraction approaches. By differentiating between diverse SMS categories, including promotional content, this approach intends to improve smishing detection efficiency and reliability. Our research aims to considerably improve smishing detection by using a broader range of Bangla SMS types and addressing the drawbacks of previous binary classification systems.

The study's primary contributions are:
\begin{itemize}
    \item Introducing a novel multiclass classification approach to Bangla smishing detection, which allows for detailed separation between normal, spam, and promotional SMS.
    \item Proposing a hybrid feature extraction strategy that combines CNNs with BERT embeddings to improve the efficiency of models in capturing textual details.
    \item Using BERT's attention mechanism to boost smish identification accuracy by focusing on key text chunks.
\end{itemize}

\section{Literature Review}

Sandhya et al.~\cite{mishra_2020_smishing} suggested a smishing detection approach using the Naive Bayes algorithm through analysis of SMS content, while Ankit Sandhya et al.~\cite{jain_2022_a} suggested employing KNN, ETC and RF based on URL behavior analysis. 

Rubaiath et al.~\cite{rubaiatheulfath_2021_detecting} implemented a technique for identifying phishing attempts in SMS messages using NLP-based extraction of features and selection, achieving a remarkable 98.39\% accuracy by primarily targeting SMS containing phishing URLs. The utilization of a dataset comprising both legitimate and smishing messages, employing SVM and RF algorithms. The methodology and machine learning classifiers used in previous studies \cite{balim_2019_automatic, mdfarhanshahriyar_2024_phishing} have achieved a satisfactory level of accuracy in detecting phishing and SMiShing frauds. This shows their effectiveness in providing reliable solutions to protect victims from financial and psychological threats \cite{boukari_2021_machine}.

Conventional feature extraction techniques and machine learning algorithms frequently fall short of effectively capturing the nuanced features present in complex and unstructured SMS texts, which hampers phishing SMS detection. To address this limitation, integrating advanced deep learning techniques, such as CNN\cite{ramilyarullin_2021_bert,nizojamanshohan_2024_enhancing}, can significantly improve identifying phishing attempts in text messages. Followed by Phishing Detection framework merges a pre-trained transformer model, MPNet, using Bi-directional Gated Recurrent Units (GRU) and supervised ConvNets (CNN) is designed to detect complex patterns in unstructured short phishing text messages effectively by Rubaiath et al.~\cite{rubaiatheulfath_2021_hybrid}.

The "SpotSpam" study~\cite{oswald_2022_spotspam} discovered SMS spam detection by using BERT embeddings and intention analysis, addressing the limitations of static textual methods. This approach achieves 98.07\% accuracy, improving adaptability and stability in dynamic keyword environments.

H. Nam ~\cite{nguyen_2023_transformerbased} suggested a novel multi-class intrusion detection system (IDS) for the in-vehicle Controller Area Network (CAN) bus using an attention network built on transformers (TAN). The major finding of this study is that transformers could enhance intrusion detection by transferring knowledge from a source model to a target model.

This study~\cite{sultanzavrak_2023_email} introduced an email spam detection method using a blend of CNNs, and GRUs, with attention mechanisms. The technique enhances feature extraction through hierarchical convolution layers and improves detection accuracy by integrating attention mechanisms at both sentence and word levels. The cross-dataset evaluation shows that the proposed FT+HAN architecture, which integrates CNNs with attention mechanisms, surpasses existing methods as attention layers with unique text processing efficiency~\cite{vaswani_2017_attention}.

\section{Methodology}
The methodology for this study involves several crucial steps to effectively preprocess, encode, and classify Bangla SMS texts for smishing detection.

\begin{figure} [ht]
    \centering
    \includegraphics[width=1\linewidth]{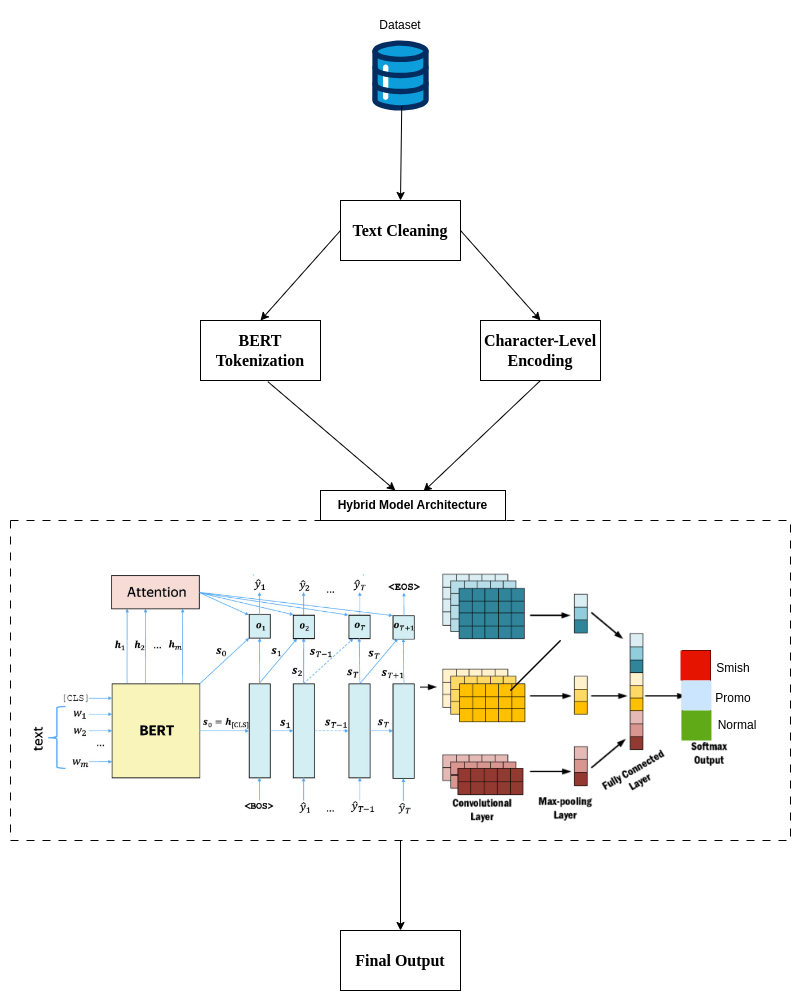}
    \caption{Methodology for Bangla SMS Smishing Detection \cite{ramilyarullin_2021_bert, xiao_2022_exploring}}
    \label{fig:methodlogy}
\end{figure}
\subsection{Dataset}
The dataset, sourced from IEEE DataPort, consists of 2,287 Bangla SMS messages categorized into three classes: Normal, Smish, and Promotional. It features two columns: 'label,' which denotes the category of each SMS (Normal, Smish, Promotional), and 'text,' containing the actual Bangla content of the SMS. The distribution of labels is 924 instances of Normal, 914 of Smish, and 449 of Promotional. Collected via an online survey and validated by cybersecurity experts, this dataset has been carefully reviewed to ensure data integrity and relevance for smishing detection. It serves as a robust foundation for developing and evaluating machine learning models aimed at improving cybersecurity measures for Bangla language\cite{vxz9-ak04-24}.

\subsection{Text Cleaning}
In this regard, we clean the Bangla SMS data, making it devoid of noise and irrelevance \cite{garg_2022_text}. Firstly, we remove all the non-alphanumeric characters that are not part of the Bangla script and allow only those characters that are relevant, i.e., Unicode range \texttt{\textbackslash u0980-\textbackslash u09FF}. Then, we normalize the text by removing extra white spaces between the words so that one space exists between two words. These basic steps make the text cleaner for proper tokenization and feature extraction.

\subsection{BERT Tokenization}
The cleaned SMS text is tokenized using the BERT tokenizer from the transformers library, a multilingual text tokenizer. This is tokenized into subword units based on the model \texttt{bert-base-multilingual-cased}\cite{devlin_2019_bert}, and then these tokens are mapped into tokens' IDs. There are produced attention masks with paddings to keep the length uniform at 128 tokens. This ensures that BERT can attend to the relevant parts of the text in the process.

\subsection{Character-Level Encoding}
We implement a custom character-level tokenizer for a more detailed analysis of the Bangla script \cite{alshehri_2022_characterlevel}. The sentences are converted into lists of integer IDs using vocabulary developed from the dataset. The encoded sequences are padded or truncated to a fixed length of 256 characters. Special tokens, like \texttt{<PAD>} and \texttt{<UNK>}, are included to manage padding and unknown characters. This lets the model catch fine-grained features in addition to word-level information coming from BERT.

\begin{equation}
  \text{encoded}_{\text{padded}}[i] = 
    \begin{cases} 
    \text{encoded}[i] & \text{for } i < E \\
    \text{char\_vocab}(\text{$<$PAD$>$}) & \text{for } i \geq E 
    \end{cases}  
\end{equation}
\text{where } $E = \min(L, \text{max\_len})$

\subsection{Model Architecture}

\subsubsection{Hybrid Model Design}
The hybrid model combines BERT for word-level embeddings and a Convolutional Neural Network (CNN) for character-level features \cite{kaur_2023_bertcnn}. BERT extracts the contextualized word representation that assists in grasping the semantic nuances of the Bangla SMS text \cite{mdmahbuburrahman_2020_bangla}. CNN processes character-level encoded text, searching for morphological variation together with local patterns using convolutional layers of different filter sizes. Outputs from BERT and CNN are concatenated to form the hybrid feature vector by collecting information from word and character levels.

\subsubsection{Attention Mechanism}
We then introduce an attention mechanism to enable the model to concentrate more on the components of the text that really matter. It selectively emphasizes those features within the SMS class that are more indicative of the SMS category for better detecting subtle cues.

\begin{figure} [!ht]
    \centering
    \includegraphics[width=1\linewidth]{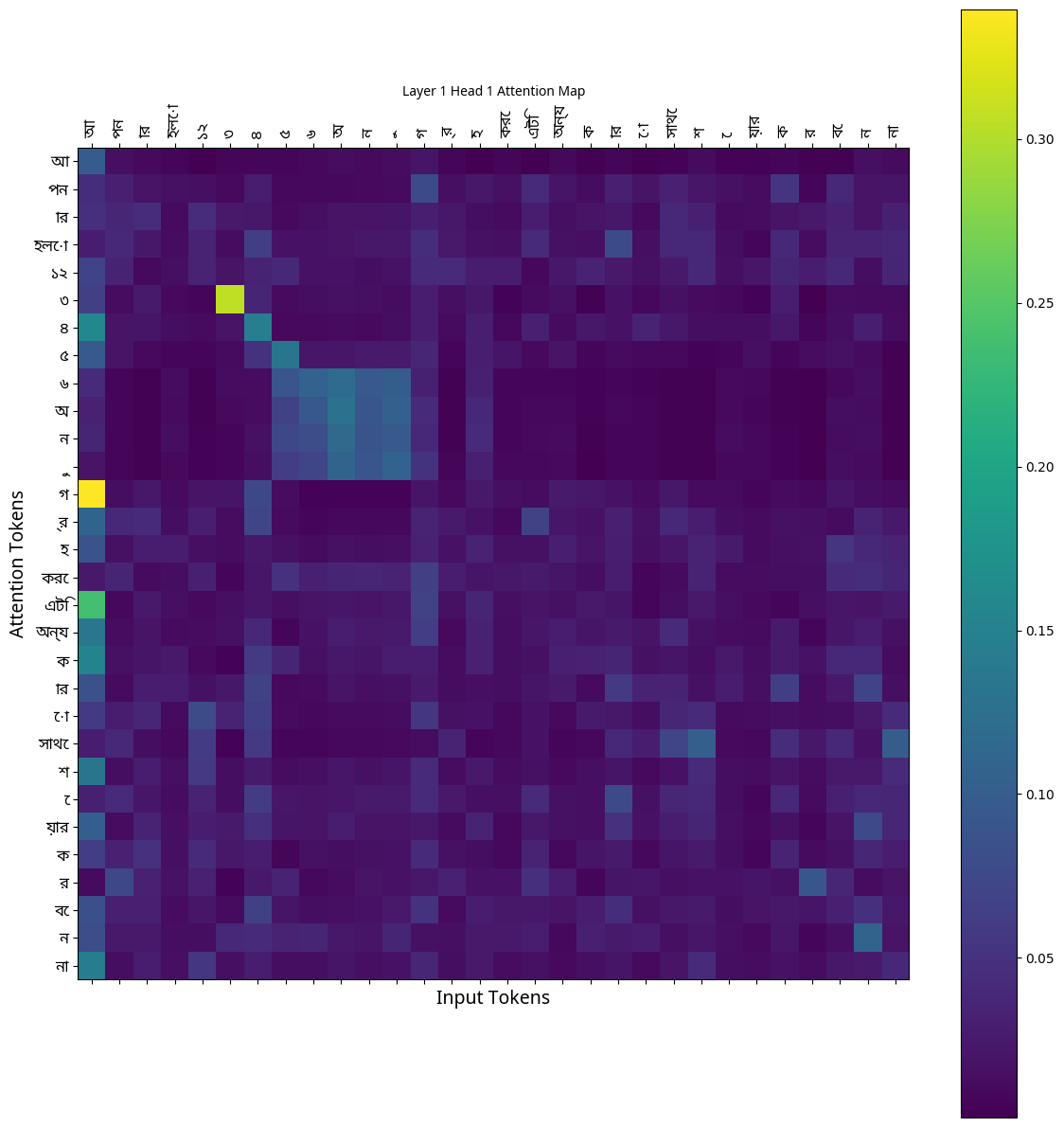}
    \caption{Attention Map for Layer 1, Head 1 in BERT Model. This heatmap visualizes the attention weights assigned by the BERT model to Bengali text input, illustrating how different tokens in the input text influence each other. The filtered Bengali tokens are represented by the x-axis and y-axis, and the color intensity reflects the degree of attention given to each token pair. This visualization helps in understanding the focus and relationships between words in the context of the smishing detection task.}
    \label{fig:attention-layer}
\end{figure}

\subsubsection{Classification Layer}
This passed Hybrid Feature vector goes through another fully connected layer and a classification layer of three categories of SMS: Smish, Normal, and Promotional. It uses the rich multilevel features drawn by the model for more informed classification decisions.

\subsection{Model Training}
Training is carried out on the preprocessed Bangla SMS dataset for several epochs using the hybrid model. The BERT model is fine-tuned for SMS classification, and the CNN layers are trained for the extraction of character-level features. Such a combined training strategy would be helpful in tapping word- and character-level information features.

\subsection{Loss Function and Optimizer}

The model utilizes the cross-entropy loss function, which is effective in classifying many class issues. The loss function can be written as follows:

\begin{equation}
    \mathcal{L} = - \sum_{i=1}^{C = 3} y_i \log(\hat{y}_i)
\end{equation}

Where the number of classes is represented by \( C \), the actual label for class \( i \) is represented by \( y_i \), and the projected probability for class \( i \) is indicated by \( \hat{y}_i \).

We used AdamW algorithm with a learning rate of \( 2 \times 10^{-5} \) for optimization. The adjustment formula for the AdamW optimizer is expressed as:

\begin{equation}
\theta_{t+1} = \theta_t - \eta \left( \frac{\hat{m}_t}{\sqrt{\hat{v}_t} + \epsilon} \right) - \eta \lambda \theta_t
\end{equation}

Where The learning rate is represented by \(\eta \), the weight decay parameter is indicated by \(\lambda \), and the first and second moments of the gradients are estimated by \(\hat{m}_t \) and \(\hat{v}_t \), respectively. A tiny constant, \(\epsilon \), prevents division by zero.

After three epochs of training the model with the loss function monitored at each step, ensures continuous improvement in classification accuracy.

\section{Results and Analysis}

\subsection{Model Performance}

The proposed Hybrid CNN-BERT model delivered strong results across all three SMS classification categories—Normal, Promotional, and Smishing—within Bangla SMS. The Precision, Recall, and F1-scores obtained were:

\begin{table}[ht]
\centering
\caption{Model Evaluation Metrics}
\begin{tabular}{lcccc}
\hline
\textbf{Category} & \textbf{Precision} & \textbf{Recall} & \textbf{F1-Score} & \textbf{Support} \\
\hline
Normal & 0.97 & 0.99 & 0.98 & 178 \\
Promo  & 0.98 & 0.96 & 0.97 & 90  \\
Smish  & 0.99 & 0.97 & 0.98 & 190 \\
\hline
\textbf{Accuracy} & \multicolumn{4}{c}{0.98 (458)} \\
\textbf{Macro Avg} & 0.98 & 0.97 & 0.98 & 458 \\
\textbf{Weighted Avg} & 0.98 & 0.98 & 0.98 & 458 \\
\hline
\end{tabular}
\label{tab:model_performance}
\end{table}

\begin{itemize}
    \item \textbf{Normal SMS}: The model attained a precision of 0.97, an almost perfect recall of 0.99, and an F1-score of 0.98, highlighting its effectiveness in accurately detecting all instances of normal SMS in the test dataset.
    \item \textbf{Promotional SMS}: The model's performance in this category was slightly lower, and has a 0.96 recall and 0.98 F1-score, indicating a minor difficulty in distinguishing between promotional and other types of SMS.
    \item \textbf{Smishing SMS}: The model performed exceptionally well in this category, achieving a 0.99 F1-score, 0.97 precision, and 0.98 recall demonstrating its capability in classifying smishing messages.
\end{itemize}

\subsubsection{General Accuracy}

\begin{equation}
\text{Accuracy} = \frac{TP + TN}{TP + TN + FP + FN}
\end{equation}

\noindent where:
\begin{itemize}
    \item \( TP \) denotes True Positives
    \item \( TN \) signifies True Negatives
    \item \( FP \) represents False Positives
    \item \( FN \) stands for False Negatives
\end{itemize}

The overall accuracy of the model was 0.98, with F1-scores, recall, and Precision each having weighted average values and macro-values of 0.98, indicating balanced performance across all categories.

\subsubsection{Confusion Matrix Analysis}
The confusion matrix highlights the hybrid CNN-BERT model's strong performance in classifying Normal and Smishing SMS, with near-perfect accuracy and minimal misclassifications. Specifically, the model correctly identified nearly all Normal SMS and flawlessly classified all Smishing SMS. However, a small number of Promotional SMS were misclassified as Smishing, indicating a slight challenge in distinguishing between these categories. Overall, while the model excels in detecting Smishing attempts, some refinement is needed to improve its precision in differentiating between Promotional and Smishing SMS.

\begin{figure} [h!]
    \centering
    \includegraphics[width=1\linewidth]{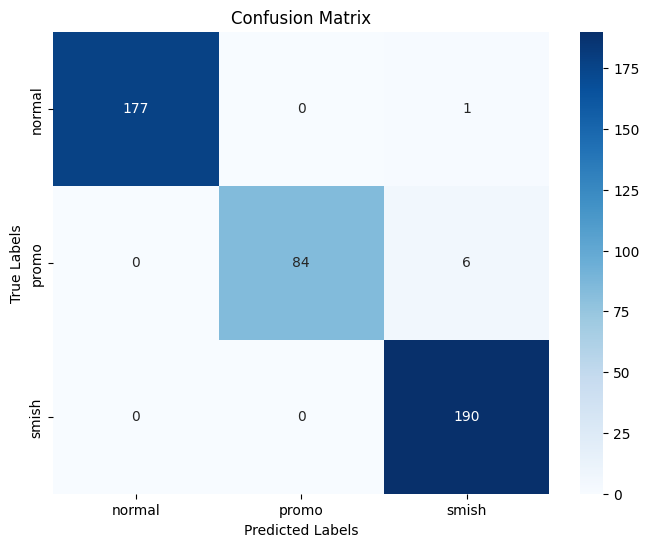}
    \caption{Confusion Matrix for Hybrid CNN-BERT Model }
    \label{fig:Conf-matrix}
\end{figure}

\subsection{Comparison with Other Models}

To determine the hybrid CNN-BERT model's efficacy, we compared its performance with several other commonly used models in text classification tasks. Random Forest, Naive Bayes, SVM, and Logistic Regression are among the models that are contrasted. Each model's performance metrics are summarized below:

\begin{table}[ht]
\centering
\caption{Performance Comparison of Various Models}
\begin{tabular}{|l|c|c|c|c|}
\hline
\textbf{Model} & \textbf{Accuracy} & \textbf{Precision} & \textbf{Recall} & \textbf{F1-Score} \\
\hline
Logistic Regression & 0.9389 & 0.9429 & 0.9389 & 0.9382 \\
\hline
SVM & 0.9432 & 0.9459 & 0.9432 & 0.9429 \\
\hline
Random Forest & 0.9534 & 0.9545 & 0.9534 & 0.9526 \\
\hline
Naive Bayes & 0.9112 & 0.9157 & 0.9112 & 0.9105 \\
\hline
CNN-BERT & 0.9847 & 0.9848 & 0.9847 & 0.9846 \\
\hline
\end{tabular}
\end{table}

\noindent
In our study, the CNN-BERT model fared better than any other model. With accuracies ranging from 0.9112 to 0.9534, Logistic Regression, SVM, Random Forest, and Naive Bayes demonstrated respectable performance; nevertheless, they were not as good as the CNN-BERT model. The CNN-BERT model showed outstanding performance in identifying Normal, Promotional, and Smishing SMS with an impressive accuracy of 0.9847, coupled with good precision, recall, and F1-scores of 0.9846.

\begin{figure} [ht]
    \centering
    \includegraphics[width=1\linewidth]{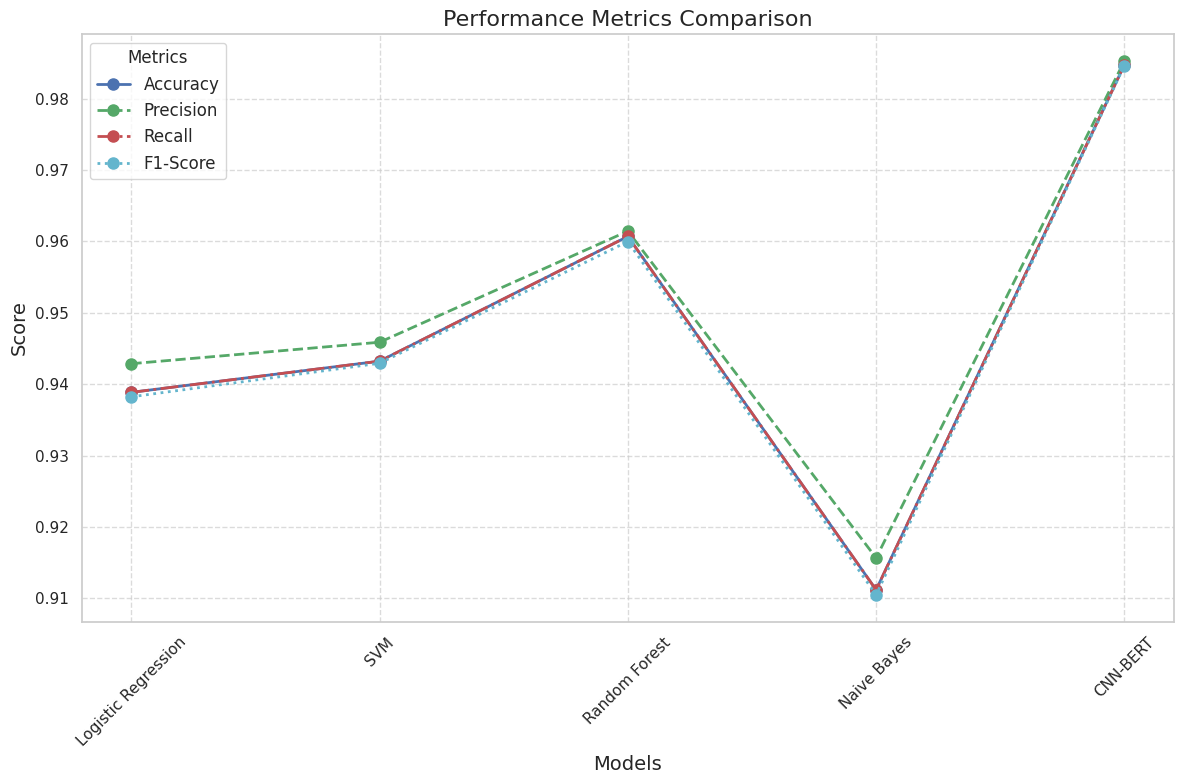}
    \caption{Comparing Machine Learning Models' Performance}
    \label{fig:comparison}
\end{figure}

The comparison illustrates that the hybrid CNN-BERT model not only surpasses traditional models like Logistic Regression, SVM, and Naive Bayes but also outperforms more complex models like Random Forest. This highlights the effectiveness of the hybrid approach in leveraging both character-level and contextual embeddings for accurate SMS classification, confirming its excellent results when used for smishing detection.

\subsection{Analysis and Discussion}

To evaluate the hybrid CNN-BERT model for smishing detection in Bangla SMS, five-fold cross-validation was employed. The dataset was split into five subsets, with each serving as a validation set once while the other four were used for training. This approach allowed us to estimate generalization accurately. The training of the model was done with the AdamW optimizer, which has a learning rate of \(2 \times 10^{-5}\) over three epochs in each fold. It was verified after training, and important metrics including F1-score, accuracy, and recall were computed. The average accuracy across all folds was 0.9852, indicating consistent performance.

The model's strong performance is due to its hybrid approach combining CNN for character-level features and BERT for contextual embeddings. This methodology effectively captures detailed textual patterns and a broader semantic context, improving the capacity of the model to recognize subtle differences and classify SMS messages accurately. It excels in minimizing false negatives, particularly for normal and smishing SMS, showing high recall and precision. However, some confusion between Promotional and Smishing SMS indicates that further refinement, such as adding more features or adjusting the training process, may be needed.

\begin{figure}
    \centering
    \includegraphics[width=1\linewidth]{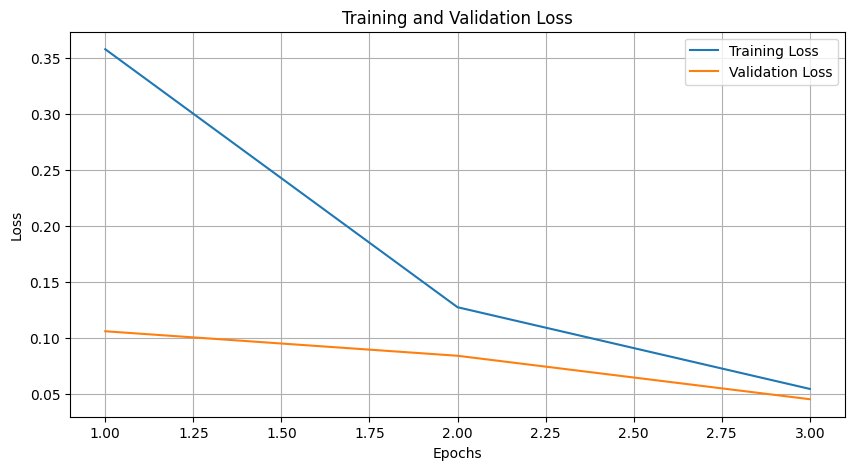}
    \caption{Loss curves for training and validation throughout the course of the three epochs.}
    \label{fig:loss}
\end{figure}
The graph of training loss versus validation loss shows a steady decline in both metrics across the three epochs. Starting with a training loss of 0.3581 and validation loss of 0.1063 in the first epoch, the losses decreased to 0.0548 and 0.0456, respectively, by the third epoch. This trend indicates effective learning and generalization, with the model fitting the training data while working effectively on the validation set in a balanced manner, without overfitting.

Overall, the hybrid CNN-BERT model shows great potential for real-world use in smishing detection, providing a robust tool for enhancing mobile security.

\section{Conclusion and Future Work}

A unique hybrid machine learning model is presented in this study, for detecting Bangla smishing text, combining BERT's contextual embeddings with character-level features extracted by Convolutional Neural Networks (CNNs). Our model demonstrates a significant advancement in SMS classification by addressing the complexities inherent in multi-class scenarios, specifically distinguishing between Normal, Promotional, and Smishing SMS.

Based on the outcomes of the experiment, the hybrid CNN-BERT model performs well in each of the three areas. With an overall accuracy of 98.47\%, It outperforms more established models, including Logistic Regression, SVM, Random Forest, and Naive Bayes regarding different evaluation metrics. The integration of BERT with CNN permits the model to record semantic context and fine-grained textual details, thereby improving smishing detection accuracy.

The successful implementation of our approach highlights its effectiveness in overcoming the limitations of previous binary classification methods and provides a more nuanced solution to SMS categorization. The model's high performance, especially in detecting smishing messages, underlines its potential as a robust tool for enhancing mobile security against smishing attacks.

Future work could involve several key areas to enhance the research. Firstly, implementing the hybrid CNN-BERT model in real-time applications or as part of a broader security framework could offer practical insights into its performance and usability. This could include integrating the model into mobile security apps or SMS filtering systems to assess its effectiveness in real-world scenarios. Additionally, further research might explore alternative hybrid models that incorporate different deep learning architectures or feature extraction techniques. Such investigations could lead to improvements in classification accuracy and robustness, potentially resulting in more effective smishing detection systems.

\bibliographystyle{unsrt}
\bibliography{Smish}

\end{document}